\documentclass[letterpaper]{article} 

\usepackage{aaai25}  
\usepackage{algorithm}
\usepackage{algpseudocode}
\usepackage{amsmath}
\usepackage{amsthm}
\usepackage{booktabs}  
\usepackage{caption} 
\usepackage{cleveref}
\usepackage{courier}  
\usepackage{graphicx} 
\usepackage{helvet}  
\usepackage[hyphens]{url}  
\usepackage{multirow}
\usepackage{natbib}  
\usepackage{times}  
\usepackage{threeparttable}  

\allowdisplaybreaks
\DeclareMathOperator*{\argmax}{arg\,max}

\Crefname{prop}{Proposition}{Propositions}

\title{Observation Adaptation via Annealed Importance Resampling for Partially Observable Markov Decision Processes}
\author{
    Yunuo Zhang \textsuperscript{\rm 1}, 
    Baiting Luo \textsuperscript{\rm 1}, 
    Ayan Mukhopadhyay \textsuperscript{\rm 1}, 
    Abhishek Dubey \textsuperscript{\rm 1}
}
\affiliations{
    \textsuperscript{\rm 1} Vanderbilt University
}

\begin{document}
    \maketitle

    \begin{abstract}
        Partially observable Markov decision processes (POMDPs) are a general mathematical model for sequential decision-making in stochastic environments under state uncertainty. POMDPs are often solved \textit{online}, which enables the algorithm to adapt to new information in real time. Online solvers typically use bootstrap particle filters based on importance resampling for updating the belief distribution. Since directly sampling from the ideal state distribution given the latest observation and previous state is infeasible, particle filters approximate the posterior belief distribution by propagating states and adjusting weights through prediction and resampling steps. However, in practice, the importance resampling technique often leads to particle degeneracy and sample impoverishment when the state transition model poorly aligns with the posterior belief distribution, especially when the received observation is highly informative. We propose an approach that constructs a sequence of bridge distributions between the state-transition and optimal distributions through iterative Monte Carlo steps, better accommodating noisy observations in online POMDP solvers. Our algorithm demonstrates significantly superior performance compared to state-of-the-art methods when evaluated across multiple challenging POMDP domains.
    \end{abstract}

    \section{Introduction} \label{sec:introduction}
    Partially observable Markov decision processes (POMDPs) provide a general mathematical framework for modeling decision-making problems under uncertainty, where the true state of the environment is not fully observable and actions have probabilistic outcomes \cite{POMDPs_Definition_Kaelbling}. These models have been successfully applied to various real-world scenarios, including time-critical UAV search and rescue operations where efficient path planning must balance computational constraints with effective decision-making \cite{Shrinking_POMCP_Zhang}. However, POMDPs face challenges to solve exactly due to the ``curse of dimensionality'' and the ``curse of history'', which make the computation of optimal policies non-scalable~\cite{POMDPs_Challenges_Papadimitriou}. To address these computational and scalability issues, online planning algorithms have emerged as a prominent approach. Instead of computing a full policy offline, online planning interleaves planning and execution by focusing computational resources on the current belief state and the immediate decisions to be made \cite{POMDPs_Planning_Ross}.

    A critical component of online planning algorithms is the belief update process where belief denotes a distribution over the possible states. At each node in the search tree, the algorithm needs to update the belief state based on simulated actions and observations. However, performing exact belief updates is computationally infeasible in large state spaces due to the high dimensionality of the belief space \cite{Large_Belief_Updates_Rodriguez}. To manage this, online planning algorithms approximate the belief state using sampling-based methods. They represent the belief state as a collection of sampled states, or particles, rather than as explicit probability distributions \cite{POMCP_Silver}. During the planning process, belief updates are performed by propagating these particles through the state-transition and observation models, using methods such as direct sampling and sequential importance sampling \cite{Particle_Filtering_Doucet}. 

    However, these approximation methods face a significant issue: the variance of their estimations increases exponentially with the search depth. This variance escalation is primarily due to the accumulation of uncertainty from observations, which provide imperfect information about the true current state. As the search depth increases, the divergence between the sampled belief and the target posterior can grow substantially, potentially limiting the effectiveness of these planning algorithms, especially in deeper searches where precise belief representation is crucial.

    Recognizing this challenge, we propose an approach, AIROAS (Annealed Importance Resampling for Observation Adaptation Search), a novel
    approach for POMDP planning that combines tree search with particle-based belief representation and annealed importance resampling. It aims to reduce the increasing variance by gradually refining the sampled states. Instead of relying solely on samples drawn from the state-transition distribution, we implement an annealed importance sampling process that incrementally transforms these samples. The goal is to shift the distribution of these samples closer to the ideal state distribution, which would be conditioned on the latest observation and the previous state. While directly sampling from this ideal distribution is impossible, our approach approximates it by using Markov Chain Monte-Carlo approximation. By aligning the sampled states more closely with the most probable current states given the available information, we improve the quality of the belief representation. This enhancement leads to more accurate state estimations, particularly at greater search depths. 

    \section{Background} \label{sec:background}
    \subsection{POMDPs} \label{sec:POMDPs}
        Partially observable Markov decision processes (POMDPs) are a framework for modeling decision-making under uncertainty, defined by the tuple $(S, A, T, R, \Omega, O)$. Here, $S$ is the set of states, $A$ the set of actions, $T(s' | s, a)$ the transition function, $R(s, a)$ the reward function, $\Omega$ the set of observations, and $O(o | s', a)$ the observation function. In POMDPs, at time t, the agent maintains a belief state $b_t$, a probability distribution over states. This distribution is updated after action $a_t$ and observation $o_{t+1}$ using the equation: $b_{t+1}(s') = \eta O(o_{t+1} | s', a_t) \sum_{s \in S} T(s' | s, a_t) b_t(s)$, where $\eta$ is a normalizing factor. This belief update process enables the agent to estimate its current state despite partial observability, facilitating informed decision-making in uncertain environments.

    \subsection{Particle Filters} \label{sec:particle_filters}
        Particle filters ~\cite{Particle_Filtering_Arulampalam}, also known as sequential Monte Carlo (SMC) methods, are computational algorithms used to estimate the state of a dynamic system when it is observed through noisy measurements. In the context of Partially observable Markov decision processes (POMDPs), particle filters offer an efficient way to approximate the belief state by maintaining a set of weighted samples (particles) that represent the probability distribution over possible states. The widely-known bootstrap particle filter algorithm ~\cite{Bootstrap_Particle_Filtering_Gordon} works by recursively propagating these particles through the state transition model and updating their weights based on new observations. In POMDPs, at each time step $t$, the particle filter aims to approximate the optimal posterior distribution: 

        \begin{equation}
            \label{eq:target_distribution}
            p(s_t | o_t, s_{t-1}, a) \propto p(o_t | s_t, a) \cdot p(s_t | s_{t-1}, a)
        \end{equation}
        where $s_t$ is the current state, $o_t$ is the current observation, $s_{t-1}$ is the previous state, and $a$ is the action taken.

        A crucial step in bootstrap particle filter is resampling \cite{Resampling_Kitagawa}, which addresses the problem of particle degeneracy. Over time, some particles may have negligible weights, leading to poor representation of the state space. Resampling involves randomly drawing particles with replacement from the current set, with probabilities proportional to their weights, and then resetting all weights to $1/N$, where $N$ is the number of particles. However overly frequent resampling can lead to sample impoverishment, where the particle set loses diversity and fails to adequately represent the full state space. To determine when to resample, the Effective Sample Size (ESS) \cite{ESS_Ronald} is often used:
        \begin{equation} \label{eq:ESS}
            ESS = \frac{1}{\sum_{i=1}^N (w_i^n)^2}
        \end{equation}
        where $w_i^n$ are the normalized weights. When the ESS falls below a predefined threshold (typically $N/2$), resampling is triggered. 

        Although the bootstrap particle filter is relatively easy to implement, its accuracy can be significantly compromised when there is a substantial discrepancy between the optimal posterior distribution described in equation~(\ref{eq:target_distribution}) and the state-transition distribution. This limitation becomes particularly pronounced in scenarios where the one-step received observation is highly informative about the true state, leading to a sharply peaked posterior distribution. In such cases, the bootstrap filter's reliance on the state-transition model for proposal generation may lead to particles being proposed in regions where the posterior has significant mass but the proposal density is low, resulting in many particles having negligible weights. Consequently, this can exacerbate the problem of sample degeneracy, necessitating more frequent resampling and potentially leading to sample impoverishment.

    \subsection{Annealed Importance Sampling}\label{sec:annealed_importance_sampling}
        Importance Sampling (IS) is a technique for estimating properties of a target distribution by sampling from a different, easier-to-sample proposal distribution and reweighting the samples \cite{Importance_Sampling_Kloek}. However, its major limitation lies in its inefficiency when the proposal distribution differs significantly from the target distribution --- in such cases, only a few samples that happen to fall in the high-probability regions of the target distribution receive high importance weights, while most samples have negligible weights. This leads to high variance in the estimates and poor sample efficiency.

        Annealed Importance Sampling (AIS) constructs a sequence of intermediate distributions that gradually bridge the gap between the proposal and target distributions \cite{Annealed_Importance_Sampling_Neal}. Given a target density $p(x)$ and a proposal density $q(x)$, AIS defines a sequence of intermediate distributions:
        \begin{equation}
        \pi_k(x) = p(x)^{\beta_k}q(x)^{(1-\beta_k)}
        \end{equation}
        where $0 = \beta_0 \leq \cdots \leq \beta_K = 1$ represents a sequence of inverse temperatures. The method proceeds by first drawing an initial sample $x_0 \sim \pi_0(x)$, then evolving this sample through a series of transition kernels:
        
        \begin{equation} \label{eq:transition_kernel}
        x_k \sim T_k(x_k|x_{k-1}) \text{ for } k = 1,\ldots,K
        \end{equation}
        where each $T_k$ is constructed to leave $\pi_k$ invariant. The final importance weight is computed as the product of ratios between successive distributions:
        
        \begin{equation}
        w = \frac{\pi_1(x_1)}{\pi_0(x_1)} \cdots \frac{\pi_K(x_K)}{\pi_{K-1}(x_K)}
        \end{equation}
        This gradual transition through intermediate distributions helps AIS overcome the limitations of standard importance sampling by maintaining better overlap between successive distributions, resulting in more reliable estimates.

    \subsection{Related Work} \label{sec:related_work}
        Monte Carlo tree search (MCTS) has demonstrated success in solving large POMDPs online. POMCP \cite{POMCP_Silver} pioneered this approach by combining a UCT-based tree search with particle filtering for belief updates, but faces challenges with continuous observation spaces. DESPOT and its variant AR-DESPOT \cite{ARDESPOT_Ye} improve upon POMCP by focusing the search on a fixed set of scenarios and using dual bounds to guide exploration, making it more robust for large discrete problems. POMCPOW \cite{POMCPOW_Sunberg} extends these ideas to continuous state-action-observation spaces by incorporating progressive widening and weighted particle filtering. 

        A common thread among these approaches is their reliance on particle filtering techniques, specifically bootstrap particle filtering or Sequential Importance Resampling (SIR), to update belief states. While these filtering methods are computationally efficient, they can sometimes lead to particle degeneracy issues, especially in continuous observation spaces or when observations are unlikely under the current belief.

        Most recently, AdaOPS \cite{AdaOPS_Wu} attempts to address this challenge using KLD-sampling, which dynamically adapts the number of particles based on the Kullback-Leibler divergence between the true and approximated distributions. However, KLD-sampling exhibits two major limitations: First, it requires discretizing the state space into bins for divergence calculation, which becomes computationally prohibitive especially in high-dimensional spaces \cite{Disadvantage_KLD_Li}. Second, and more critically, while it adjusts particle quantities, it does not modify the particle values themselves, leading to potential underestimation of distribution variance and failure to capture multimodal aspects of the target distribution. In contrast, Annealed Importance Sampling (AIS) offers several compelling advantages: it constructs a sequence of intermediate distributions that gradually bridge the prior and posterior distributions, effectively maintaining particle diversity while preventing degeneracy. This annealing process enables particles to adaptively migrate toward regions of high posterior probability, making it particularly effective when dealing with concentrated observation likelihoods or significant disparities between prior and posterior distributions. The gradual transition through intermediate distributions allows AIS to better capture the full structure of multimodal distributions and provide more accurate representations of distribution tails compared to KLD-sampling approaches.

        Previous work has explored adding tempering iterations to particle filters, where the optimal posterior distribution is constructed adaptively through a sequence of Monte Carlo steps. For example, \citet{Tempered_Particle_Filtering_Johansen} developed a block-tempered particle filter that uses bridge distributions to gradually adapt particle values, while \cite{Tempered_Particle_Filtering_Herbst} proposed a tempered particle filter that sequentially reduces inflated measurement error variance in Dynamic Stochastic General Equilibrium (DSGE) models. However, these tempering approaches have been primarily studied in the context of state estimation and system identification, but not yet explored within the POMDP planning literature. To our knowledge, we are the first to investigate using bridge distributions to connect target and proposal distributions in particle filtering specifically for POMDP planning, combining the strengths of both sequential Monte Carlo methods and POMDP planning algorithms.

    \section{Approach} \label{sec:approach}
\begin{figure*}
            \includegraphics[width=\linewidth]{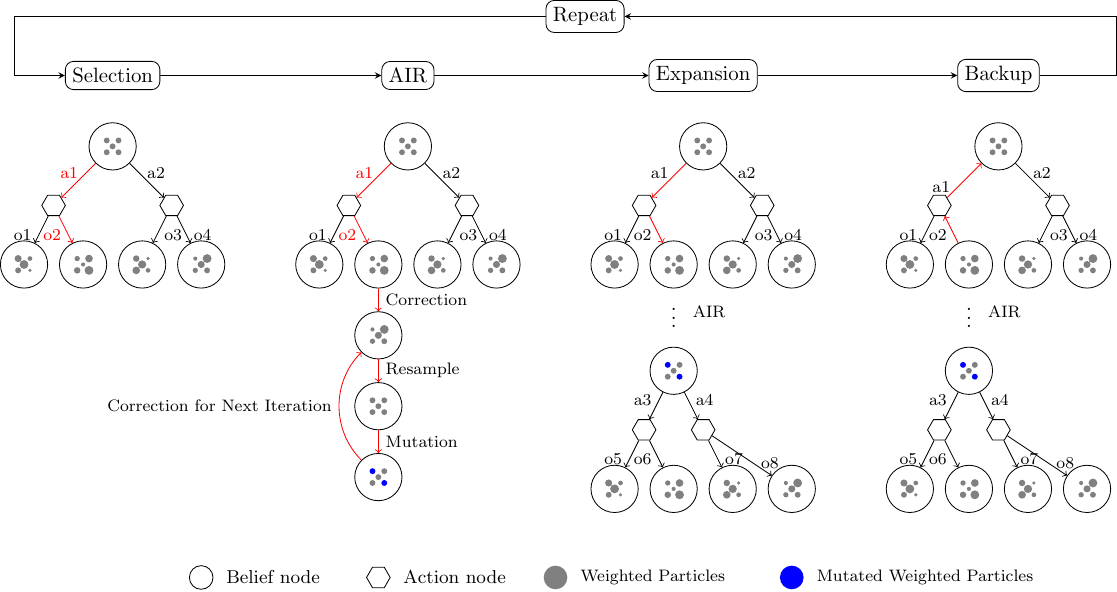}
            \caption{Process of AIROAS Tree Search. AIROAS constructs a search tree that alternates between belief nodes and action nodes. AIR represents Annealed Importance Resampling.}
            \label{fig:tree_process}
        \end{figure*}

    In this section, we present AIROAS (Annealed Importance Resampling for Observation Adaptation Search), a novel approach for POMDP planning that combines tree search with particle-based belief representation and annealed importance resampling. The complete procedure is detailed in~\cref{alg:AIROAS} and ~\cref{fig:tree_process}. AIROAS constructs a search tree that alternates between belief nodes and action nodes. Each belief node approximates the optimal posterior distribution using a set of weighted particles, as illustrated in~\cref{fig:tree_process}, and maintains both upper and lower bounds of the optimal value. Using a sequence of bridging distributions, the algorithm refines the value bounds and adjusts particle states and weights to better approximate the optimal posterior distribution described in equation~(\ref{eq:target_distribution}). At each timestep, starting from the current belief $b_0$ as the root node, AIROAS expands the belief tree by exploring various paths from the initial states (Line 3 in \cref{alg:AIROAS}). A key characteristic of this structure is that sibling nodes share identical particle states but maintain distinct weight distributions. This configuration persists until annealed importance resampling is applied to mutate the particle values, as we will discuss below.

    \subsection{Selection} \label{sec:search}

        \begin{algorithm}[!ht]
    \caption{AIROAS}
    \begin{algorithmic}[1]
    \Require initial belief $\bar{b}_0$, maximum depth MAX\_DEPTH
    \While{Time Allowed and $l(\bar{b}_0) < u(\bar{b}_0)$}
        \State $\bar{b} \gets \bar{b}_0$
        \While{depth$(\bar{b}) < \text{MAX\_DEPTH}$}
            \If{$\bar{b}$ is a leaf node}
                \State $\bar{b} \gets \text{Annealed Importance Resampling}(\bar{b})$ 
                \State ExpandAndBackup$(\bar{b})$
            \EndIf
            \State $\bar{b} \gets $ Selection$(\bar{b})$
        \EndWhile
        \If{depth$(\bar{b}) \geq \text{MAX\_DEPTH}$}
            \State $u(\bar{b}) \gets l(\bar{b})$
        \EndIf
    \EndWhile
    \State \Return $\arg\max_{a \in \mathcal{A}} l(\bar{b}_0, a)$
    \end{algorithmic}
    \label{alg:AIROAS}
\end{algorithm}

        \begin{algorithm}[!ht]
    \caption{Selection$(\bar{b})$}
    \begin{algorithmic}[1]
        \State $a^* = \argmax_{a \in A} u(\bar{b}, a)$
        \State $o^* \leftarrow \arg\max_{o \in O_{\bar{b},a}} \hat{p}(o|\bar{b},a)\text{EU}(\tau(\bar{b},a^*,o))$
        \State $\bar{b} \leftarrow \tau(\bar{b}, a^*, o^*)$
    \end{algorithmic}
    \label{alg:Selection}
\end{algorithm}

        Similar to \cite{AdaOPS_Wu, ARDESPOT_Ye, HSVI_Smith}, during online planning, AIROAS maintains both upper and lower bounds, $u(\bar{b}, a)$ and $l(\bar{b}, a)$, for each action node $a$ at a belief node $\bar{b}$. These bounds estimate the optimal value that can be achieved by taking action $a$ at belief $\bar{b}$. The action selection follows an optimistic strategy - at each belief node $\bar{b}$, AIROAS chooses the action $a^*$ that maximizes the upper bound(Line 1 in \cref{alg:Selection}):

        \begin{equation}
            a^* = \argmax_{a \in A} u(\bar{b}, a)
        \end{equation}
        
        After selecting an action, we need to choose which observation branch to explore next. This choice is guided by the probability-weighted excess uncertainty (EU) criterion~\cite{ARDESPOT_Ye, HSVI_Smith} (Line 2 in \cref{alg:Selection}). For each possible observation $o$ after taking action $a^*$, we compute:
        
        \begin{equation}
            \hat{p}(o | \bar{b}, a^*) \cdot EU(\tau(\bar{b}, a^*, o))
        \end{equation}
        where $\hat{p}(o | \bar{b}, a^*)$ is an estimation of the probability $p(o | \bar{b}, a^*)$ i.e estimates the probability of receiving observation $o$ after taking action $a^*$ at belief $\bar{b}$, $\tau(\bar{b}, a^*, o)$ represents the updated belief after executing $a^*$ and receiving observation $o$, and $EU(\cdot)$ measures the excess uncertainty at a belief node ~\cite{AdaOPS_Wu}. The excess uncertainty quantifies the degree of value uncertainty at each belief node:
        
        \begin{equation}
            EU(\bar{b}) = [u(\bar{b}) - l(\bar{b})] - \frac{\xi[u(\bar{b}_0) - l(\bar{b}_0)]}{\gamma^{d(\bar{b})}}
        \end{equation}
        where $u(\bar{b})$ and $l(\bar{b})$ are the upper and lower bounds at belief $\bar{b}$, $\bar{b}_0$ is the root belief, $d(\bar{b})$ denotes the depth of belief $\bar{b}$ in the tree, $\gamma$ is the discount factor, and $\xi \in (0,1)$ controls the target uncertainty reduction at the root. The observation $o^*$ that maximizes the weighted excess uncertainty is selected for exploration, effectively focusing the search on belief states with high uncertainty and high probability of being reached \cite{AdaOPS_Wu}.

    \subsection{Expand And Backup} \label{sec:expand_and_backup}
        When encountering a leaf node, we first perform annealed importance resampling to adjust the particle states and weights (detailed in the next section). Then, AIROAS expands this leaf node by creating child nodes for each possible action $a$. For each action node, the expansion process involves propagating the parent belief node's particles forward using a simulator $\mathcal{G}$ that generates state transitions (Line 6 in \cref{alg:ExpandAndBackup}):
        \[
        (s', o, r) \sim \mathcal{G}(s, a)
        \]
        \begin{algorithm}[!ht]
    \caption{ExpandAndBackup$(\bar{b})$}
    \begin{algorithmic}[1]
        \For{$a$ $\in$ $\mathcal{A}$}
            \State Create Action Node ($\bar{b}, a$)
            \State $b^\prime$ $\gets$ Empty Particles Set
            \State $\mathcal{O}$ $\gets$ Empty Observation Set
            \For{s $\in$ $\bar{b}$}
                \State $(s\prime, o, r) \sim \mathcal{G}(s, a)$
                \State $\mathcal{O}$ $\gets$ $\mathcal{O}$ $\cup$ o
                \State weight($s\prime$) $\gets$ 1
                \State $b^\prime$ $\gets$ $b^\prime$ $\cup$ $s\prime$
            \EndFor
            \For{o $\in$ $\mathcal{O}$}
                \State $\Tilde{b}$ $\gets$ DeepCopy($b^\prime$)
                \For{s $\in$ $\Tilde{b}$}
                    \State weight(s) = $p(o | s, a)$ $\cdot$ weight(s)
                \EndFor
                \State Create Belief Node $\Tilde{b}$
            \EndFor
        \EndFor
        \State BackUp($\bar{b}$)
    \end{algorithmic}
    \label{alg:ExpandAndBackup}
\end{algorithm}

        The expansion continues by creating new belief nodes for each unique observation encountered during particle propagation. Specifically, after propagating each particle through $\mathcal{G}$, we obtain a set of updated particle states, initially assigning each particle a weight of 1. For each observation $o$ generated during this process, we create a new belief node where the particle weights are updated according to the observation likelihood $p(o|s,a)$ --- the probability of receiving observation $o$ given the particle state $s$ and action $a$. This results in a tree structure where each action node branches into multiple belief nodes based on possible observations, with each belief node containing weighted particles that represent the updated belief state.

        After expanding belief nodes, we perform a backup operation to update the bounds of their ancestor nodes. Similar as \cite{ARDESPOT_Ye} and \cite{AdaOPS_Wu}, this backup process is implemented by recursively applying the Bellman equation, which calculates the optimal value of a belief state in two parts. First, we consider the expected immediate reward for taking an action in the current belief state, calculated by looking at each possible state, weighing its reward by how likely we think we are in that state according to our current belief. Second, we consider the long-term value by looking at all possible observations we might receive after taking an action. For each possible observation, we calculate how likely we are to see that observation, determine what our new belief state would be after seeing it, consider the value of that resulting belief state, and weight this value by the probability of getting that observation. The total value is the sum of these immediate and future components, and we choose the action that maximizes this total value. In practice, since we can't compute exact values over continuous states and observations, we approximate this calculation using discrete sums over our particle-based belief representation.

    \subsection{Annealed Importance Resampling} \label{sec:annealed_importance_resampling}
        As discussed above, the belief of each belief node is generated using a state-transition function from its parent belief node. Suppose the leaf node represents belief at timestep t and its parent belief node represents belief at timestep t-1, while our simulator $\mathcal{G}$ implements the state-transition function $p(s_t | s_{t-1}, a)$ during expansion, our ultimate objective is to have the final particle distribution approximate the optimal posterior distribution $p(s_t | o_t, s_{t-1}, a)$ referenced in \cref{eq:target_distribution}.

        \begin{algorithm}[!ht]
    \caption{Annealed Importance Resampling($\bar{b}$)}
    \begin{algorithmic}[1]
    \Require $\bar{b}$ from current belief node, sequence of tempering parameters $0 = \beta_0 \leq \cdots \leq \beta_K = 1$, target inefficiency ratio $r^*$, transition kernels $T_1(s, s'), \ldots, T_K(s, s')$
    \For{$\beta_k$ in $\{\beta_0, \ldots, \beta_K\}$}
        \If{$k == 0$}
            \State Continue
        \EndIf
        \State weights($\bar{b}$) $\gets$ Update\_Weights($\bar{b}$, $\beta_k$, $\beta_{k-1}$)
        \If{In\_eff($\bar{b}$) $\leq r^*$}
            \State break
        \Else
            \State $r^*$ $\gets$ In\_eff($\bar{b}$)
        \EndIf
        \State Resample $\bar{b}$
        \State $\bar{b}$ $\gets$ Mutation($\bar{b}$)
    \EndFor
    \end{algorithmic}
    \label{alg:annealed_importance_resampling}
\end{algorithm}

        The primary challenge in this context stems from the impossibility of directly sampling from the optimal posterior distribution. Previous work \cite{POMCP_Silver, ARDESPOT_Ye, POMCPOW_Sunberg, AdaOPS_Wu} approximates sampling from this optimal posterior distribution using Sequential Importance Resampling, treating the state-transition distribution as proposal distribution and the optimal posterior distribution as target distribution. However, this approach suffers from particle degeneracy issues, particularly when observations are highly informative about the true state, leading to a sharply peaked posterior distribution and a significant divergence between the optimal posterior and the proposal distribution derived from the current belief \cite{AdaOPS_Wu}.

        To address the issue, Annealed Importance Sampling \cite{Annealed_Importance_Sampling_Neal} introduced in Background offers an effective solution by constructing intermediate bridging distributions between the proposal distribution and target distribution, gradually guiding particles toward high-probability regions of the target distribution by constructing a sequence of intermediate distributions parameterized by tempering parameters $0 = \beta_0 \leq \beta_1 \leq \cdots \leq \beta_K = 1$.

        Building upon these principles, we integrate Annealed Importance Resampling (AIR), which extends AIS with a resampling step, into our AIROAS algorithm to address the particle degeneracy issue inherent in POMDP planning. Our approach adapts this concept to the POMDP planning context by designing intermediate distributions that account for observations. By gradually bridging the proposal and target distributions, particles are adaptively reweighted and resampled, allowing them to migrate toward regions of high posterior probability. This gradual transition through intermediate distributions not only addresses degeneracy but also enables our approach to capture the multimodal characteristics of the optimal posterior distribution more effectively.

        These tempering parameters are generated using a sigmoid function applied to a linearly spaced sequence between $10^{-3}$ and 1. Specifically, for a sequence $\{x_1, x_2, \cdots x_K\}$ linearly spaced between $10^{-3}$ and 1, each tempering parameter $\beta_i$ is computed as:
        \begin{equation} \label{eq:betas}
            \beta_i = \frac{1}{1 + e^{-10(x_i - 0.5)}}
        \end{equation}
        This sigmoid transformation ensures a smooth progression of $\beta$ values from near zero to one, with denser sampling in the middle range.

        The intermediate distributions are characterized by their probability densities:
        \begin{equation} \label{eq:bridging_distributions}
            \pi_k = p(s_t | o_t, s_{t-1}, a)^{\beta_k}p(s_t | s_{t-1}, a)^{(1-\beta_k)}
        \end{equation}
        When $k = K$, this formulation exactly matches the optimal posterior distribution. By progressively transforming particles sampled from the prior distribution $p(s_t | s_{t-1}, a)$ through the sequence $\pi_1, \pi_2, \ldots, \pi_K$, we can effectively approximate particle states drawn from the optimal posterior distribution.

        The overall procedure of annealed importance resampling is described in~\cref{alg:annealed_importance_resampling}.

    \subsection{Weights} \label{sec:update_weights}
        To update weights of the particles at iteration $k$ and timestep t, for every $\beta_k$ in the sequence $\{\beta_0, \ldots, \beta_K\}$, we transform the particle approximation from $\pi_{k-1}$ to $\pi_k$ by recalculating the weights. Assume the particle set at iteration $k$ and timestep t is $S_{tk}$, for each particle $s_{tk}^j$ in the particle set $S_{tk}$ at iteration $k$, the weight is updated as:
        \begin{equation}\label{eq:weight}
            w_{tk}^j \propto \frac{p(o_t \mid s_{tk}^j, a_t)^{\beta_k}}{p(o_t \mid s_{tk}^j, a_t)^{\beta_{k-1}}} \cdot w_{t(k-1)}^j
        \end{equation}

        \begin{proof}
    At iteration k-1 and timestep t, for particle $s_{t(k-1)}^j$, we have computed its weight as: 
    \begin{equation*}
        w_{t(k-1)}^j = \frac{\pi_1(s_{t1}^j)}{\pi_0(s_{t1}^j)} \cdots \frac{\pi_{k-1}(s_{t(k-1)}^j)}{\pi_{k-2}(s_{t(k-1)}^j)}
    \end{equation*}
    Thus
    \begin{equation*}
        w_{tk}^j = \frac{\pi_k(s_{tk}^j)}{\pi_{k-1}(s_{tk}^j)} \cdot w_{t(k-1)}^j
    \end{equation*}
    From \cref{eq:bridging_distributions} and \cref{eq:target_distribution}, we know:
    \begin{align*}
        \pi_k(s_{tk}^j) &= p(s_{tk}^j | o_t, s_{t-1}^j, a_t)^{\beta_k}p(s_{tk}^j | s_{t-1}^j, a_t)^{(1-\beta_k)} \\
         &\propto p(o_t | s_{tk}^j, a_t)^{\beta_k} p(s_{tk}^j | s_{t-1}^j, a_t)^{\beta_k}p(s_{tk}^j | s_{t-1}^j, a_t)^{(1-\beta_k)} \\
         &\propto p(o_t | s_{tk}^j, a_t)^{\beta_k} p(s_{tk}^j | s_{t-1}^j, a_t)
    \end{align*}
    Similarly, we can get:
    \begin{equation*}
        \pi_{k-1}(s_{tk}^j) \propto p(o_t | s_{tk}^j, a_t)^{\beta_{k-1}} p(s_{tk}^j | s_{t-1}^j, a_t)
    \end{equation*}
    where $o_t$ and $a$ is the observation and action leading to this belief node. Thus we have
    \begin{align*}
        w_{tk}^j &= \frac{\pi_k(s_{tk}^j)}{\pi_{k-1}(s_{tk}^j)} \cdot w_{t(k-1)}^j \\
            &\propto \frac{p(o_t | s_{tk}^j, a_t)^{\beta_k} p(s_{tk}^j | s_{t-1}^j, a_t)}{p(o_t | s_{tk}^j, a_t)^{\beta_{k-1}} p(s_{tk}^j | s_{t-1}^j, a_t)} \cdot w_{t(k-1)}^j \\
            &\propto \frac{p(o_t | s_{tk}^j, a_t)^{\beta_k}}{p(o_t | s_{tk}^j, a_t)^{\beta_{k-1}}} \cdot w_{t(k-1)}^j
    \end{align*}
\end{proof}

        This update reflects the incremental adjustment of weights based on the annealing parameter $\beta_k$. Specifically, at iteration $k$, the importance weight of each particle $s_{tk}^j$ is updated by multiplying its previous weight $w_{t(k-1)}^j$ with an incremental weight. This incremental weight is computed as the ratio of two observation likelihood terms: the observation density function raised to the current tempering parameter $\beta_k$ in the numerator, and the same function raised to the previous tempering parameter $\beta_{k-1}$ in the denominator. Specifically, the incremental weight is $\frac{p(o_t | s_{tk}^j, a_t)^{\beta_k}}{p(o_t | s_{tk}^j, a_t)^{\beta_{k-1}}}$, where $o_t$ is the observation received at timestep $t$ and $a_t$ is the action taken.

        After updating the particle weights, we calculate their inefficiency score in a similar way as \cite{Tempered_Particle_Filtering_Herbst}. For a particle set containing $M$ particles, the inefficiency score is computed as:
        \begin{equation} \label{eq:in_eff}
            \text{In\_eff}(\bar{b}_{tk}) = \frac{1}{M} \sum_{j=1}^M (\frac{w_{tk}^j}{\frac{1}{M}\sum_{j=1}^Mw_{tk}^j})^2
        \end{equation}
        This score reflects the variance in particle weights. As in \cite{Tempered_Particle_Filtering_Herbst} and specified in \cref{alg:annealed_importance_resampling}, we maintain a pre-defined target inefficiency ratio $r^*$. If the inefficiency score exceeds $r^*$, indicating that the variance of the resulting weights remains high, we continue adjusting the particle states and weights. Otherwise, we exit the current iteration.

    \subsection{Mutation} \label{sec:mutation}
        For each particle $s_{t(k-1)}^j$, the algorithm employs a Markov transition kernel $T_k(s_{k-1}^j, s_k^j)$ to evolve the state from time $k-1$ to $k$. This transition kernel is constructed to preserve $\pi_k$ as its invariant distribution, as specified in \cref{eq:transition_kernel}.

        \begin{algorithm}[!ht]
    \caption{Mutation($\bar{b}$)}
    \begin{algorithmic}[1]
    \For{particle $s_{t(k-1)}^j$ in $\bar{b}$}
        \State $T_k(s_{k-1}^j, \cdot)$ $\gets$ $\mathcal{N}(s_{t(k-1)}^j, \sigma^2 \cdot I)$
        \State $s_{tk}^j \sim T_k(s_{k-1}^j, \cdot)$
        \State Compute the acceptance probability p
        \State $s_{tk}^j$ $\gets$ $s_{tk}^j$ for probability p, otherwise $s_{tk}^j$ $\gets$ $s_{t(k-1)}^j$
    \EndFor
    \end{algorithmic}
    \label{alg:mutation}
\end{algorithm}

        The transition mechanism implements a Metropolis-Hastings kernel~\cite{Metropolis_Hastings_Hastings} with a multivariate Gaussian proposal distribution. Given a state vector $s_{t(k-1)}^j$ that can be decomposed into components $s_{t(k-1)}^j = <x, y, \cdots>$, the proposal distribution is centered at $s_{t(k-1)}^j$ with covariance structure. The forward proposal distribution i.e the transition kernel generates new states according to:
        \begin{equation*}
            s_{tk}^j \sim \mathcal{N}(s_{t(k-1)}^j, \sigma^2 \cdot I)
        \end{equation*}
        where $I$ denotes the identity matrix and $\sigma^2$ is a scaling parameter calibrated proportionally to the L1 distance between the current state $s_{t(k-1)}^j$ and the observation $o_t$ in the state space. Then we define its reverse transition kernel $\Tilde{T_k}(s_k^j)$ as:
        \begin{equation*}
            \mathcal{N}(s_{tk}^j, \sigma^2 \cdot I)
        \end{equation*}
        
        The acceptance rate shown in \cref{alg:mutation} is computed as:
        \begin{equation} \label{eq:acceptance_rate}
            p_{accept} = \min \left\{1,  \frac{T_k(s_{t(k-1)}^j, s_{tk}^j) \cdot p(o_t | s_{tk}^j, a_t)^{\beta_k}}{\Tilde{T_k}(s_{tk}^j, s_{t(k-1)}^j) \cdot p(o_t | s_{t(k-1)}^j, a_t)^{\beta_k})}\right\}
        \end{equation}
        
        Then we update the particle state:
        \begin{equation}
            s_{tk}^j = 
            \begin{cases}
                s_{tk}^j \text{with prob. } p_{\text{accept}} \\
                s_{t(k-1)}^j \text{with prob. } 1 - p_{\text{accept}}
            \end{cases}
        \end{equation}
        
        The iterative process continues until either the particle approximation converges to the optimal posterior distribution, or the variance of the particle weights becomes less than or equal to the threshold $r^*$.

    \section{Experiments} \label{sec:experiments}
    \begin{table*}[!htbp]
    \caption{Performance Comparison}
    \label{tab:performance}
    \centering
    \begin{tabular}{lcccccc}
    \toprule
                    & LD ($\alpha=0.5$) & LD ($\alpha=1.0$)  & Tag      & Laser Tag  & RS(11, 11)   & RS(15,15) \\
    \midrule
    $|\mathcal{S}|$ & $\infty$   &   $\infty$     & 870      & 4830  & 247,808    & 7,372,800         \\             
    $|\mathcal{A}|$ & 3          &   3     & 5        & 5      & 20   & 16             \\              
    $|\mathcal{O}|$ & $\infty$   &   $\infty$     & 30       & $\sim 1.5 \times 10^6$ & 3  & 3                 \\
    \midrule
    POMCP    & $0.357 \pm 0.22$  & $0.691 \pm 0.41$     & $-18.530 \pm 0.40$   & $-18.112 \pm 0.18$ & $12.252 \pm 0.44$ & $8.207 \pm 0.53$  \\
    ARDESPOT & $-1.24 \pm 0.25$  & $0.760 \pm 0.41$     & $-13.704 \pm 0.71$    & $-15.982 \pm 0.64$ & $18.113 \pm 0.64$ & $16.111 \pm 0.66$    \\
    POMCPOW  & $1.555 \pm 0.21$  & $1.468 \pm 0.49$     & $-16.783 \pm 0.42$     & $-15.263 \pm 0.40$ & $13.521 \pm 0.67$ & $9.141 \pm 0.44$    \\
    AdaOPS   & $2.231 \pm 0.13$  & $2.762 \pm 0.27$     & $-9.920 \pm 0.70$    & $-13.427 \pm 0.67$ & $21.904 \pm 0.59$ & $19.091 \pm 0.57$    \\
    \midrule
    AIROAS   & $\mathbf{2.303 \pm 0.43}$ & $\mathbf{3.102 \pm 0.36}$ & $\mathbf{-8.451 \pm 0.75}$ & $\mathbf{-12.931 \pm 1.22}$ & $\mathbf{22.872 \pm 0.51}$ & $\mathbf{20.246 \pm 0.50}$ \\
    \bottomrule
    \end{tabular}
    \begin{tablenotes}
    \small
    \item Note: LD($\alpha$) stands for Light Dark with step size $\alpha$. RS(n,m) stands for the Rock Sample with $n \times n$ map and $m$ rocks. $\infty$ means continuous state (or observation) space. The results represent the average discounted return and its standard error of mean (SEM) for each method and each domain. Higher discounter return indicates better performance.
    \end{tablenotes}
\end{table*}

    In this section, we evaluate our method on several domains and conduct an ablation study to demonstrate the contribution of each component.

    \subsection{Baseline Approaches} \label{sec:baselines}
        We evaluate our approach against four state-of-the-art baselines: POMCP \cite{POMCP_Silver}, ARDESPOT \cite{ARDESPOT_Ye}, POMCPOW \cite{POMCPOW_Sunberg}, and AdaOPS \cite{AdaOPS_Wu} on four domains: Light Dark (LD), Tag, Laser Tag, and Rock Sample (RS) \cite{Light_Dark_Platt,POMDPs_Framework_Egorov,HSVI_Smith}. Detailed descriptions of these domains are provided in \cref{sec:domains}. For ARDESPOT and AdaOPS, we employ different initialization strategies: in Light Dark, Tag, and Laser Tag domains, we use domain-specific independent bounds based on maximum and minimum achievable discounted rewards, while for RockSample, bounds are initialized using heuristics. For POMCPOW, we optimize the maxUCB parameter---which governs action selection at each node---by testing values in \{1.0, 10.0, 20.0\} and selecting the best-performing configuration for each domain. The performance comparison across all domains is presented in \cref{tab:performance}. Our experimental results for the baselines may differ from previously published results due to modifications in implementation code, differences in computing infrastructure, and alternative bound initialization methods. Specifically, while \cite{AdaOPS_Wu} uses heuristics for bound initialization in Light Dark, Tag, and Laser Tag, we initialize bounds based on maximum and minimum achievable discounted rewards. However, we maintain consistent initialization methods across all approaches within each domain to ensure fair comparison. Complete details of hyperparameter selection for each domain are provided in \cref{sec:hyperparameters_selection}.

    \subsection{Experiment Settings} \label{sec:experiment_settings}
        The main configurations for all baselines are detailed in Baseline Approaches. For fair comparison, we ensure AIROAS's maximum allowed time per decision matches those of comparable baselines with similar parameters. While POMCP and POMCPOW use iterations rather than time limits, we maintain consistency by keeping the number of iterations identical during testing. All these parameters are documented in \cref{sec:hyperparameters_selection}. AIROAS uses the same bound initialization methods as ARDESPOT and AdaOPS. For AIROAS-specific configurations, we generate 100 tempering parameters $\beta_k$ ranging from 0 to 1 using the sigmoid transformation described in \cref{eq:betas}. We evaluate AIROAS using different target inefficiency ratios $r^* \in \{2.0, 3.0, 5.0, 10.0\}$ and select the best-performing value for each domain. All experiments were conducted on a computer equipped with an Intel(R) Core(TM) i9-14900KS processor with 32 CPUs.

        \subsection{Results}
            The average discounted return and its standard error of mean (SEM) are presented in \cref{tab:performance}. In all these domains, AIROAS outperforms the other solvers with suitable target inefficiency ratio and tempering parameters.
            
            In the Light Dark Problem, we evaluate performance with step sizes $\alpha=0.5$ and $\alpha=1.0$. The state consists of position coordinates and termination status (terminated or not), while observations provide noisy measurements of the current position. For particle state mutation, we generate a Gaussian distribution over the position components, where the variance is proportional to the distance between the state's position and the received observation. Through this mutation strategy in bridging distributions, combined with weight adjustments, we observe significant mitigation of the sample impoverishment problem, leading to improved performance.
            
            In the Tag problem, the state consists of the robot's position, target's position, and tagging status. While the agent's position is fully observable, the target's position is only observed when both actors occupy the same cell. For particle state mutation, we generate a Gaussian distribution exclusively over the target's position, as the agent's position is known. Given this limited observability structure, we did not anticipate significant performance improvements from Annealed Importance Resampling. The observed improvements may be attributed to the compact state and observation spaces, where the mutation step effectively explores different states within this limited domain.
            
            The Laser Tag problem represents a more challenging variant of the Tag problem. The state space comprises the agent's position, target's position, and terminal signal, with neither position directly observable. The observation consists of noisy range readings from laser sensors in eight different directions. For particle state mutation, we construct a multivariate Gaussian distribution over both agent and target positions. Given these high-dimensional observations, which induce multimodal distributions over the state space, we anticipate Annealed Importance Sampling to demonstrate significant advantages in this domain.
            
            In the RockSample problem, the state space comprises the robot's position and the quality (good or bad) of each rock. With scenarios containing 11 or 15 rocks, this results in an extremely high-dimensional state space with a large number of states. In contrast, the observation structure is simple, providing information about only one rock's quality at a time (or no observation). For particle state mutation, we focus specifically on mutating the quality of the rock for which the observation provides information, rather than mutating the entire state vector.

        \subsection{Benefits of Annealed Importance Resampling}
            \begin{figure}
                \centering
                \includegraphics[width=\linewidth]{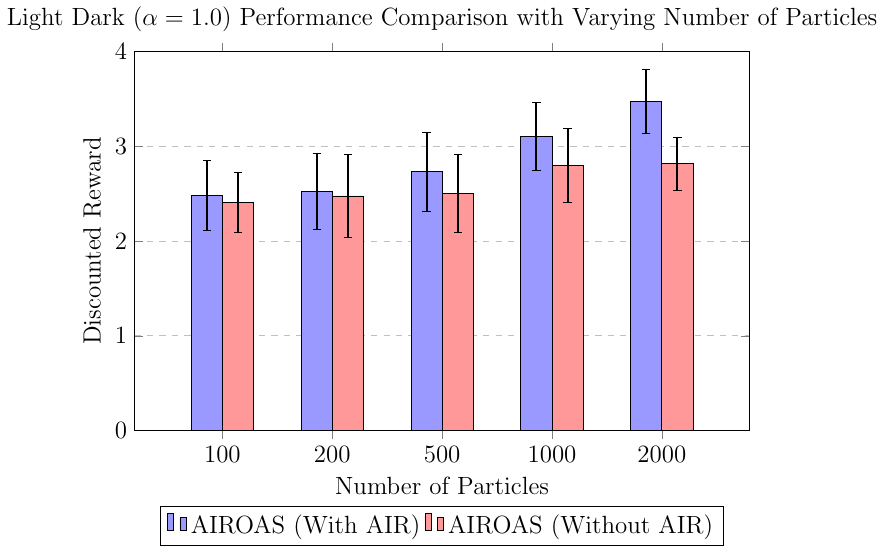}
                \caption{Performance comparison between AIROAS variants with and without Annealed Importance Resampling (AIR) on the Light Dark domain ($\alpha = 1.0$). The performance gap widens in favor of AIR as the number of particles increases.}
                \label{fig:ablation}
            \end{figure}
            
            We conduct an ablation study to evaluate the impact of Annealed Importance Resampling (AIR) on our approach. Figure \cref{fig:ablation} presents a comparative analysis between AIROAS with and without AIR on the Light Dark domain, using a step size of $\alpha = 1.0$. To ensure a comprehensive evaluation, we vary the number of particles from the set $\{100, 200, 500, 1000, 2000\}$. The results demonstrate that the performance advantage of incorporating AIR becomes more pronounced as the number of particles increases. This suggests that AIR's particle state and weight mutation strategy becomes increasingly effective with larger particle populations, while maintaining all other components of the algorithm unchanged.

    \section{Conclusion} \label{sec:conclusion}
    We presented AIROAS, a novel online POMDP solver that leverages Annealed Importance Resampling to better handle observation uncertainty. By constructing a sequence of bridge distributions between the state-transition and optimal distributions, our approach effectively addresses the particle degeneracy problem common in traditional particle filtering methods. Empirical evaluations across diverse POMDP domains demonstrate AIROAS's consistent performance advantages over state-of-the-art baselines, with ablation studies confirming that these benefits scale with the number of particles. These findings indicate that AIROAS represents a significant advancement in online POMDP solving, particularly for domains with complex observation spaces or highly informative observations that create sharply peaked posterior distributions.
    
    While our current approach uses sigmoid transformation for generating tempering parameters, future work could explore two key improvements: adaptive scheduling of tempering parameters that dynamically adjusts based on belief uncertainty, and parallel implementation of the annealing process to reduce computational overhead. Additionally, incorporating temporal abstraction techniques would enable AIROAS to efficiently handle high-dimensional continuous action spaces through learned macro-actions, extending its applicability to more complex robotic control and autonomous driving domains \cite{LMAP_Luo}. These enhancements could further improve AIROAS's efficiency and effectiveness in solving complex POMDPs across both discrete and continuous action spaces.

    \section{Acknowledgements} \label{sec:acknowledgements}
    This material is based upon work sponsored by the National Science Foundation (NSF) under Grant CNS-2238815 and by the Defense Advanced Research Projects Agency (DARPA) under the Assured Neuro Symbolic Learning and Reasoning program. Results presented in this paper were obtained using the Chameleon testbed supported by the National Science Foundation. Any opinions, findings, conclusions, or recommendations expressed in this material are those of the authors and do not necessarily reflect the views of the NSF or the DARPA.

    \bibliography{citations}

    \clearpage
    \appendix
    \setcounter{secnumdepth}{2}
    \section*{Appendix}
\section{Experiment} 
    \subsection{Domains}\label{sec:domains}
        We use the following four environments to validate our approach:
        \begin{enumerate}
            \item \textbf{Light Dark}:
            The Light Dark ~\cite{Light_Dark_Platt} is a one-dimensional continuous-state problem where an agent must navigate to a target area while dealing with state-dependent observation noise. The key challenge lies in the fact that observations, which are noisy measurements of the agent's position, become more accurate as the agent moves toward a ``light'' region and less accurate in ``darker'' regions. Starting from an initial position on one side of the environment, the agent can move left, right, or declare goal completion. We increase the problem's complexity by varying the step size $\alpha$, where larger steps make precise positioning more challenging while smaller steps increase the number of actions required for information gathering.
            
            \item \textbf{Tag}:
            The Tag problem ~\cite{POMDPs_Framework_Egorov} involves an agent that aims to tag an opponent by occupying the same grid cell and executing a tag action. The opponent moves stochastically according to a fixed policy, attempting to move away from the agent with probability $p$ while staying in the same cell with probability $1-p$. The state space consists of both agent position and target position and a tag status indicator, while the action space allows the agent to move in four cardinal directions or perform a tag action. The agent can always observe its own position but can only observe the opponent's position when both are in the same grid cell.
            
            \item \textbf{Laser Tag}:
            The LaserTag problem ~\cite{POMDPs_Framework_Egorov}, a more challenging variant of Tag, involves an agent attempting to tag an escaping target by occupying the same grid cell and executing a tag action. The state space consists of both agent position and target position and a tag status indicator, while the action space allows the agent to move in four cardinal directions or perform a tag action. Unlike Tag where the agent knows its own position, in LaserTag the agent initially has no knowledge of either its own or the target's position, and must rely on sensor information from the environment to infer both positions.
            
            \item \textbf{RockSample}:
            The RockSample problem ~\cite{HSVI_Smith} models a robot's exploration task where it must navigate through a grid environment to collect valuable rocks while avoiding bad ones before reaching an exit area. The key challenge lies in the uncertainty about rock qualities, which can only be determined through noisy sensor readings that decrease in accuracy exponentially with distance from the rock. The state space consists of the robot's position and the status of rocks (good or bad), while the action space includes four movement actions (up, down, left, right), a sampling action to collect rocks, and $K$ sensing actions to check rock status. The robot receives perfect information about its position but noisy observations about rock qualities through a sensor with efficiency parameter $\alpha$, with no observations during movement or sampling actions.
        \end{enumerate}

\section{Hyperparameters} \label{sec:hyperparameters_selection}
    \begin{table*}
    \centering
    \caption{Hyperparameters Selected}
    \label{tab:hyperparameters}
    \begin{tabular}{lcccccccc}
    \hline
    & & LD($\alpha=0.5$) & LD($\alpha=1.0$) & Tag & Laser Tag & RS(11, 11) & RS(15, 15) \\
    \hline
    \multirow{2}{*}{POMCP} & Iterations & 20000 & 20000 & 20000 & 20000  & 20000 & 20000 \\
    & Max\_Depth & 100 & 100  & 100 & 100 & 100 & 100 \\
    \hline
    \multirow{4}{*}{ARDESPOT} & K & 30 & 30 & 300 & 300 & 100 & 100 \\
    & $\lambda$ & 0.1 & 0.1 & 0.01 & 0.01 & 0.0 & 0.0 \\
    & T\_max & 5.0 & 5.0 & 5.0 & 5.0 & 5.0 & 5.0 \\
    & bounds & (-11.0, 11.0) & (-11.0, 11.0) & (-20.0, 0.0) & (-20.0, 0.0) & (FA, MDP) & (FA, MDP) \\
    \hline
    \multirow{5}{*}{POMCPOW} & c & 10 & 10 & 10 & 10 & 10 & 10 \\
    & $\alpha_{\mathcal{O}}$ & 0.03 & 0.03 & 0.03 & 0.03 & 1.0 & 1.0 \\
    & $k_{\mathcal{O}}$ & 4.0 & 4.0 & 4.0 & 4.0 & 1.0 & 1.0 \\
    & Iterations & 20000 & 20000 & 20000 & 20000  & 20000 & 20000 \\
    & Max\_Depth & 100 & 100  & 100 & 100 & 100 & 100 \\
    \hline
    \multirow{4}{*}{AdaOPS} & $m_{min}$ & 10 & 10 & 10 & 10 & 100 & 100  \\
    & $\delta$ & 1.0 & 1.0 & 0.1 & 0.1 & 0.1 & 0.1 \\
    & T\_max & 5.0 & 5.0 & 5.0 & 5.0 & 5.0 & 5.0 \\
    & bounds & (-11.0, 11.0) & (-11.0, 11.0) & (-20.0, 0.0) & (-20.0, 0.0) & (FA, MDP) & (FA, MDP) \\
    \hline
    \multirow{3}{*}{AIROAS} & $r^*$ & 5.0 & 3.0 & 2.0 & 2.0 & 10.0 & 2.0 \\
    & T\_max & 5.0 & 5.0 & 5.0 & 5.0 & 5.0 & 5.0 \\
    & bounds & (-11.0, 11.0) & (-11.0, 11.0) & (-20.0, 0.0) & (-20.0, 0.0) & (FA, MDP) & (FA, MDP) \\
    \hline
    \end{tabular}
    \begin{tablenotes}
    \small
    \item Note: LD($\alpha$) stands for Light Dark with step size $\alpha$. RS(n,m) stands for the Rock Sample with $n \times n$ map and $m$ rocks.
    \item Note: For bounds, (a,b) represents independent bound that lower value is set to a and upper value is set to b. For RockSample domain, we use heuristics to initialize the lower and upper bound values. FA means Fixed Action Policy and MDP means MDP approximation for upper bound.
    \end{tablenotes}
\end{table*}

    The hyperparameters for AIROAS and baseline algorithms are presented in \cref{tab:hyperparameters}. To ensure fair comparison, we maintain consistent initialization methods for the lower and upper bounds across ARDESPOT, AdaOPS, and AIROAS within each domain. Some of our results differ from previously published findings due to our use of different bound initialization methods in certain domains.

    For the RockSample domain, we utilize the heuristics described in \cite{AdaOPS_Wu}. Specifically, we initialize the lower bound using a Fixed Action Policy, which selects the same action regardless of observations received. The upper bound is initialized using MDP approximation. For the Light Dark, Tag, and Laser Tag domains, we employ domain-specific initialization values that differ from those reported in \cite{AdaOPS_Wu}. While this leads to some discrepancies with previously reported results, our comparison remains fair as we consistently apply the same initialization method across all approaches within each domain.
    
    For AIROAS evaluation, we tested different target inefficiency ratios ($r^*$) for each domain, as discussed in the main paper. We explored $r^*$ values in the set \{2.0, 5.0, 10.0, 20.0\} and report the best-performing results. The optimal $r^*$ value for each domain is documented in \cref{tab:hyperparameters}.
        
\end{document}